\documentclass[sigconf]{acmart}

\usepackage{booktabs} 
\usepackage{layouts}
\usepackage{tabularx}
\usepackage{multirow}
\usepackage{amsmath}
\usepackage{amssymb}
\usepackage{multicol}
\usepackage[utf8x]{inputenc} 

\copyrightyear{2020} 
\acmYear{2020} 
\setcopyright{acmcopyright}\acmConference[GECCO '20]{Genetic and Evolutionary Computation Conference}{July 8--12, 2020}{Cancún, Mexico}
\acmBooktitle{Genetic and Evolutionary Computation Conference (GECCO '20), July 8--12, 2020, Cancún, Mexico}
\acmPrice{15.00}
\acmDOI{10.1145/3377930.3389841}
\acmISBN{978-1-4503-7128-5/20/07}

\begin{document}
\title{Evolutionary Optimization of Deep Learning Activation~Functions}
\titlenote{G. Bingham and W. Macke contributed equally.}


\author{Garrett Bingham}
\orcid{0000-0002-2898-2582}
\affiliation{%
  \institution{The University of Texas at Austin and Cognizant Technology Solutions}
  \streetaddress{}
  \city{San Francisco} 
  \state{California} 
  \postcode{}
}
\email{bingham@cs.utexas.edu}

\author{William Macke}
\affiliation{%
  \institution{The University of Texas at Austin}
  \streetaddress{}
  \city{Austin} 
  \state{Texas} 
  \postcode{}
}
\email{wmacke@cs.utexas.edu}

\author{Risto Miikkulainen}
\affiliation{%
  \institution{The University of Texas at Austin and Cognizant Technology Solutions}
  \streetaddress{}
  \city{San Francisco} 
  \state{California}}
\email{risto@cs.utexas.edu}

\renewcommand{\shortauthors}{G. Bingham, W. Macke, and R. Miikkulainen}

\begin{abstract}
The choice of activation function can have a large effect on the performance of a neural network.  While there have been some attempts to hand-engineer novel activation functions, the Rectified Linear Unit (ReLU) remains the most commonly-used in practice.  This paper shows that evolutionary algorithms can discover novel activation functions that outperform ReLU.  A tree-based search space of candidate activation functions is defined and explored with mutation, crossover, and exhaustive search.  Experiments on training wide residual networks on the CIFAR-10 and CIFAR-100 image datasets show that this approach is effective.  Replacing ReLU with evolved activation functions results in statistically significant increases in network accuracy.  Optimal performance is achieved when evolution is allowed to customize activation functions to a particular task; however, these novel activation functions are shown to generalize, achieving high performance across tasks.  Evolutionary optimization of activation functions is therefore a promising new dimension of metalearning in neural networks.
\end{abstract}

%
%

\ccsdesc[500]{Computing methodologies~Neural networks}
\ccsdesc[100]{Computing methodologies~Object identification}
\ccsdesc[300]{Theory of computation~Evolutionary algorithms}

\keywords{Activation functions, evolutionary algorithms, metalearning}

\maketitle

\section{Introduction}

Together with topology, loss function, and learning rate, the choice of activation function plays a large role in determining how a neural network learns and behaves.  
An activation can be any arbitrary function that transforms the output of a layer in a neural network.
However, only a small number of activation functions are widely used in modern deep learning architectures. 
The Rectified Linear Unit, $\textrm{ReLU}(x) = \max\{x, 0\}$, is popular because it is simple and effective.  Other activation functions such as $\textrm{tanh}(x)$ and $\sigma(x) = 1 / (1 + e^{-x})$ are commonly used when it is useful to restrict the activation value within a certain range.  There have also been attempts to engineer new activation functions to have certain properties.  For example, Leaky ReLU \cite{maas2013rectifier} allows information to flow when $x < 0$.  Softplus \cite{nair2010rectified} is positive, monotonic, and smooth.  Many hand-engineered activation functions exist \cite{nwankpa2018activation}, but none have achieved widespread adoption comparable to ReLU.  Activation function design can therefore be seen as a largely untapped resource in neural network design.
 
  
This situation points to an interesting opportunity: it may be possible to optimize activation functions automatically through metalearning.  Rather than attempting to directly optimize a single model as in traditional machine learning, metalearning seeks to find better performance over a space of models, such as a collection of network architectures, hyperparameters, learning rates, or loss functions~\cite{finn2017model, finn2017meta, elsken2019neural, wistuba2019survey, feurer2015initializing, gonzalez2019improved}.  Several techniques for metalearning have been proposed, including gradient descent, Bayesian hyperparameter optimization, reinforcement learning, and evolutionary algorithms \cite{chen2019progressive, feurer2015initializing, zoph2016neural, real2019regularized}.  Of these, evolution is the most versatile and can be applied to several aspects of neural network design.

This work develops an evolutionary approach to optimize activation functions.  Activation functions are represented as trees, and novel activation functions are discovered through crossover, mutation, and exhaustive search in expressive search spaces containing billions of candidate activation functions.  The resulting functions are unlikely to be discovered manually, yet they perform well, surpassing traditional activation functions like ReLU on image classification tasks CIFAR-10 and CIFAR-100.

This paper continues with Section \ref{sec:related_work}, which presents a summary of related research in metalearning and neural network optimization, and discusses how this work furthers the field. Section \ref{sec:evolution} explains how activation functions are evolved in detail, defining the search space of candidate functions as well as the implementation of crossover and mutation.  In Section \ref{sec:experiments}, a number of search strategies are presented, and results of the experiments are given in Section \ref{sec:results}.  This work continues with discussion and potential future work in Section \ref{sec:discussion}, before concluding in Section \ref{sec:conclusion}.

\begin{figure}
    \centering
    \includegraphics[width=\linewidth]{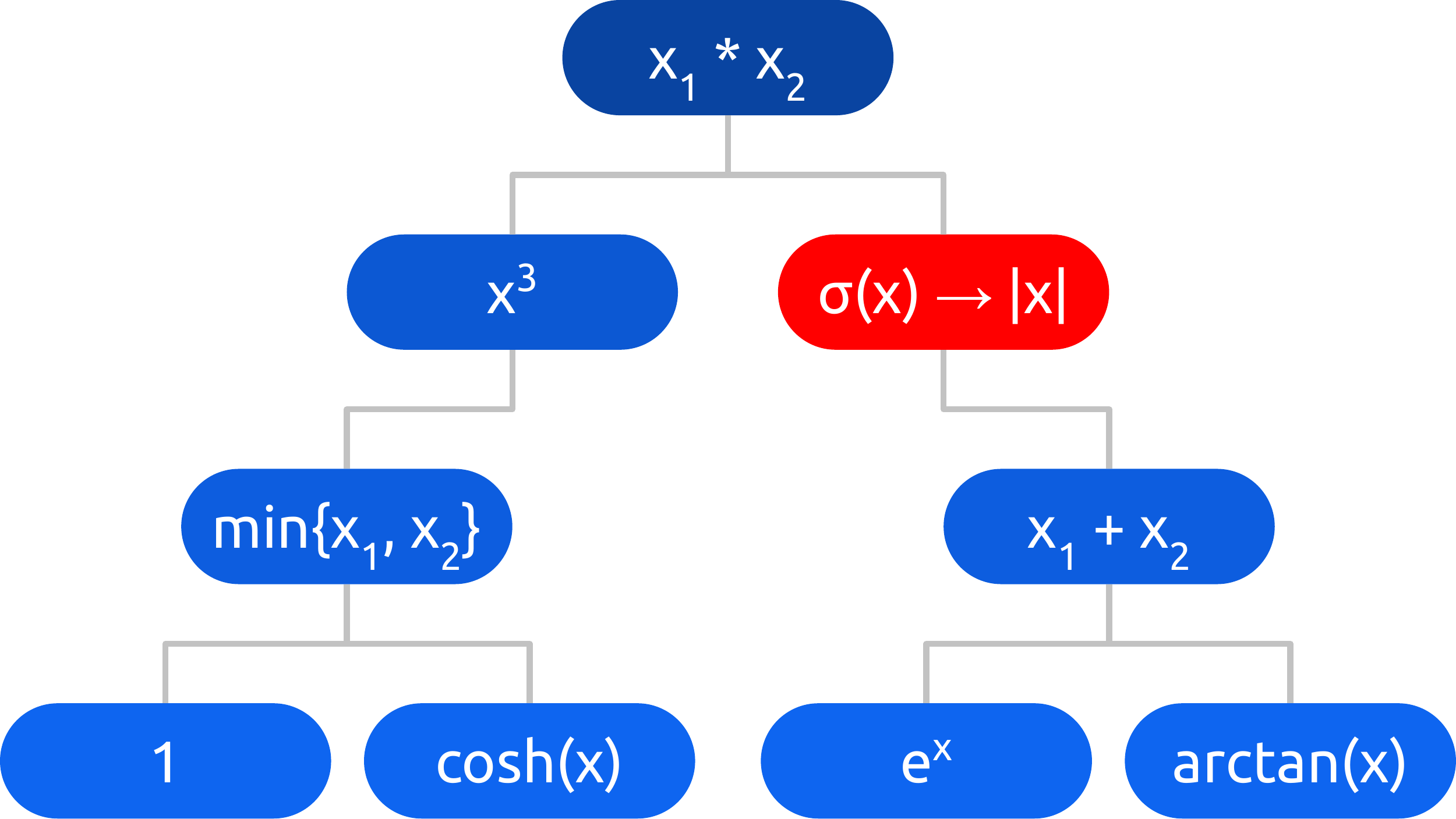}
    \caption{An example of activation function mutation.  The tree represents the activation function {\normalfont $(\min\{1, \textrm{cosh}(x)\})^3 * \sigma(e^x + \textrm{arctan}(x))$}.  One node in the tree is selected uniformly at random and replaced with another operator in the search space, also uniformly at random.  The resulting activation function is {\normalfont $(\min\{1, \textrm{cosh}(x)\})^3 * |e^x + \textrm{arctan}(x)|$}.  By introducing variability, mutation ensures evolution explores the search space sufficiently.  It prevents high-performing activation functions from overly skewing the early generations of the search process.}
    \label{fig:mutation}
\end{figure}

\section{Related Work}
\label{sec:related_work}

There has been a significant amount of recent work on metalearning with neural networks.  Neuroevolution, the optimization of neural network architectures with evolutionary algorithms, is one of the most popular approaches~\cite{elsken2019neural, wistuba2019survey}.  Real et al. \cite{real2017large} proposed a scheme where pairs of architectures are sampled, mutated, and the better of the two is trained while the other is discarded.  Xie and Yuille \cite{xie2017genetic} implemented a full genetic algorithm with crossover to search for architectures.  Real et al. \cite{real2019regularized} introduced a novel evolutionary approach called aging evolution to discover an architecture that achieved higher ImageNet classification accuracy than any previous human-designed architecture.  Beyond neuroevolution, other approaches to neural architecture search include Monte Carlo tree search to explore a space of architectures \cite{negrinho2017deeparchitect} and reinforcement learning (RL) to automatically generate architectures through value function methods \cite{baker2016designing} and policy gradient approaches \cite{zoph2016neural}.

While the above work focuses on optimizing neural network topologies, some work has also focused on optimizing other aspects of neural networks.  For instance, Gonzalez and Miikkulainen \cite{gonzalez2019improved} used a genetic algorithm to construct novel loss functions, and then optimized the coefficients of the loss functions with a covariance-matrix adaptation evolutionary strategy.  They discovered a loss function that results in faster training and higher accuracy compared to the standard cross-entropy loss.  Marchisio et al. \cite{marchisio2018methodology} introduced a method to automatically choose an activation function for each layer of a neural network, and Hagg et al. \cite{hagg2017evolving} augmented the NEAT algorithm \cite{stanley2002evolving} to simultaneously evolve per-neuron activation functions and the overall network topology.  Another example is work by Ramachandran et al.~\cite{ramachandran2017searching}, which automatically designed novel activation functions using RL.

This work expands upon previous research by introducing an evolutionary algorithm to design novel activation functions.  While Marchisio et al. \cite{marchisio2018methodology} and Hagg et al. \cite{hagg2017evolving} selected activation functions from predefined lists, this approach searches among billions of candidate activation functions.  Although Ramachandran et al. \cite{ramachandran2017searching} discovered a number of novel, high-performing activation functions with RL, they analyzed just one in depth: $x \cdot \sigma(x)$, which they call Swish.  This paper takes activation function design a step further: instead of only searching for a single activation function that performs reasonably well for most datasets and neural network architectures, it demonstrates that it is possible to evolve specialized activation functions that perform particularly well for specific datasets and neural network architectures, thus utilizing the full power of metalearning.


\begin{figure}
    \centering
    \includegraphics[width=\linewidth]{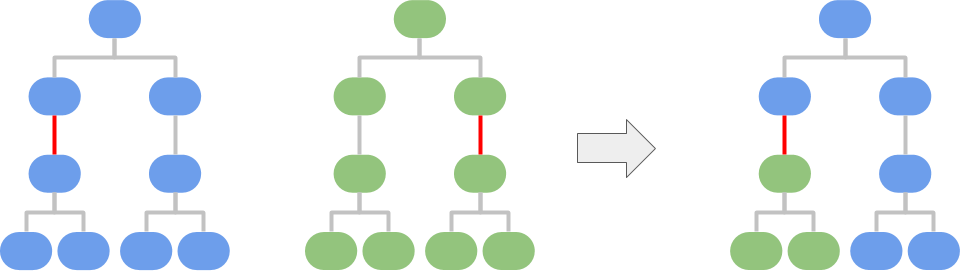}
    \caption{In crossover, two parent activation functions exchange randomly selected subtrees of equivalent depth, producing one new child activation function.  Crossover enables the best activation functions to pass on their characteristics to the rest of the population.  This mechanism is what enables evolution to discover better activation functions more quickly than random search.}
    \label{fig:crossover}
\end{figure}

\section{Evolving Activation Functions}
\label{sec:evolution}

This section presents the approach to evolving activation functions, introducing the search space, mutation and crossover implementations, and the overall evolutionary algorithm.

\subsection{Search Space}
Each activation function is represented as a tree consisting of unary and binary operators. Functions are grouped in layers such that two unary operators always feed into one binary operator.  The following operators, modified slightly from the search space of Ramachandran et al. \cite{ramachandran2017searching}, are used:
\begin{itemize}
    \item \textbf{Unary:} 0, 1, $x$, $-x$, $|x|$, $x^2$, $x^3$, $\sqrt{x}$, $e^x$, $e^{-x^2}$, \allowbreak $\log(1+e^x)$, $\log(|x + \epsilon|)$, $\sin(x)$, $\textrm{sinh}(x)$, $\textrm{arcsinh}(x)$, $\cos(x)$, $\textrm{cosh}(x)$, \allowbreak $\textrm{tanh}(x)$, $\textrm{arctanh}(x)$, $\max\{x, 0\}$, $\min\{x, 0\}$, $\sigma(x)$, $\textrm{erf}(x)$, \allowbreak $\textrm{sinc}(x)$
    \item \textbf{Binary:} $x_1 + x_2$, $x_1 - x_2$, $x_1 \cdot x_2$, $x_1 / (x_2 + \epsilon)$, $\max\{x_1, x_2\}$, $\min\{x_1, x_2\}$
\end{itemize}
Following Ramachandran et al., a ``core unit'' is an activation function that can be represented as \texttt{core\_unit = binary(unary1(x), unary2(x))}.  Let $F$ be the set of balanced core unit trees.  $S$ is then defined as a family of search spaces
\begin{equation}
S_{d\in \mathbb{N}} = \{f\in F \mid \textrm{depth}(f) = d\}.
\end{equation}
For example, $S_1$ corresponds to the set of functions that can be represented by one core unit, $S_2$ represents functions of the form: \texttt{core\_unit1(core\_unit2(x), core\_unit3(x))}, and so on. Examples of functions in $S_2$ are illustrated in Figures \ref{fig:mutation} and \ref{fig:crossover}.

  
\subsection{Mutation}
In mutation, one node in an activation function tree is selected uniformly at random.  The operator at that node is replaced with another random operator in the search space.  Unary operators are always replaced with unary operators, and binary operators with binary operators.  An example of mutation is shown in Figure~\ref{fig:mutation}.  Theoretically, mutation alone is sufficient for constructing any activation function.  However, preliminary experiments showed that crossover can increase the rate at which good activation functions are found.

\begin{figure}
    \centering
    \includegraphics[width=\linewidth]{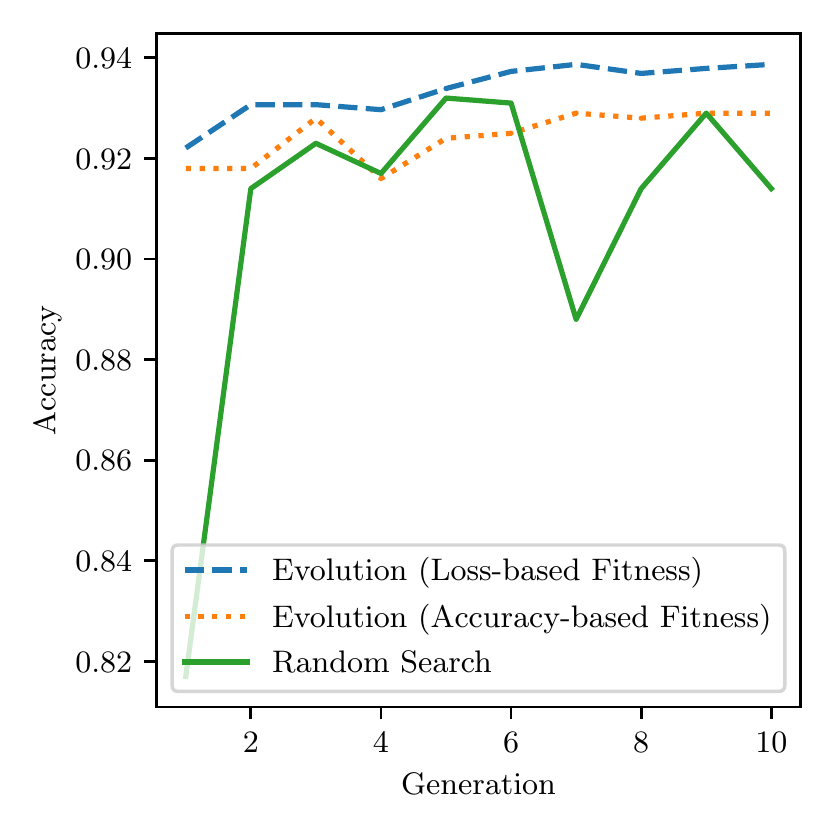}
    \caption{Top validation accuracy per generation for three search strategies in the $S_2$ search space.  There are 50 activation functions in each generation of a search.  All activation functions are trained with WRN-28-10 on CIFAR-10 for 50 epochs, and the highest validation accuracy obtained among all activation functions in a given generation is reported. Evolution with loss-based fitness finds better activation functions more quickly than evolution with accuracy-based fitness or random search. The first generation of random search is poor due to chance; since each generation is independent, the generations of random search could be arbitrarily reordered.}
    \label{fig:evolution_max}
\end{figure}

\begin{figure}
    \centering
    \includegraphics{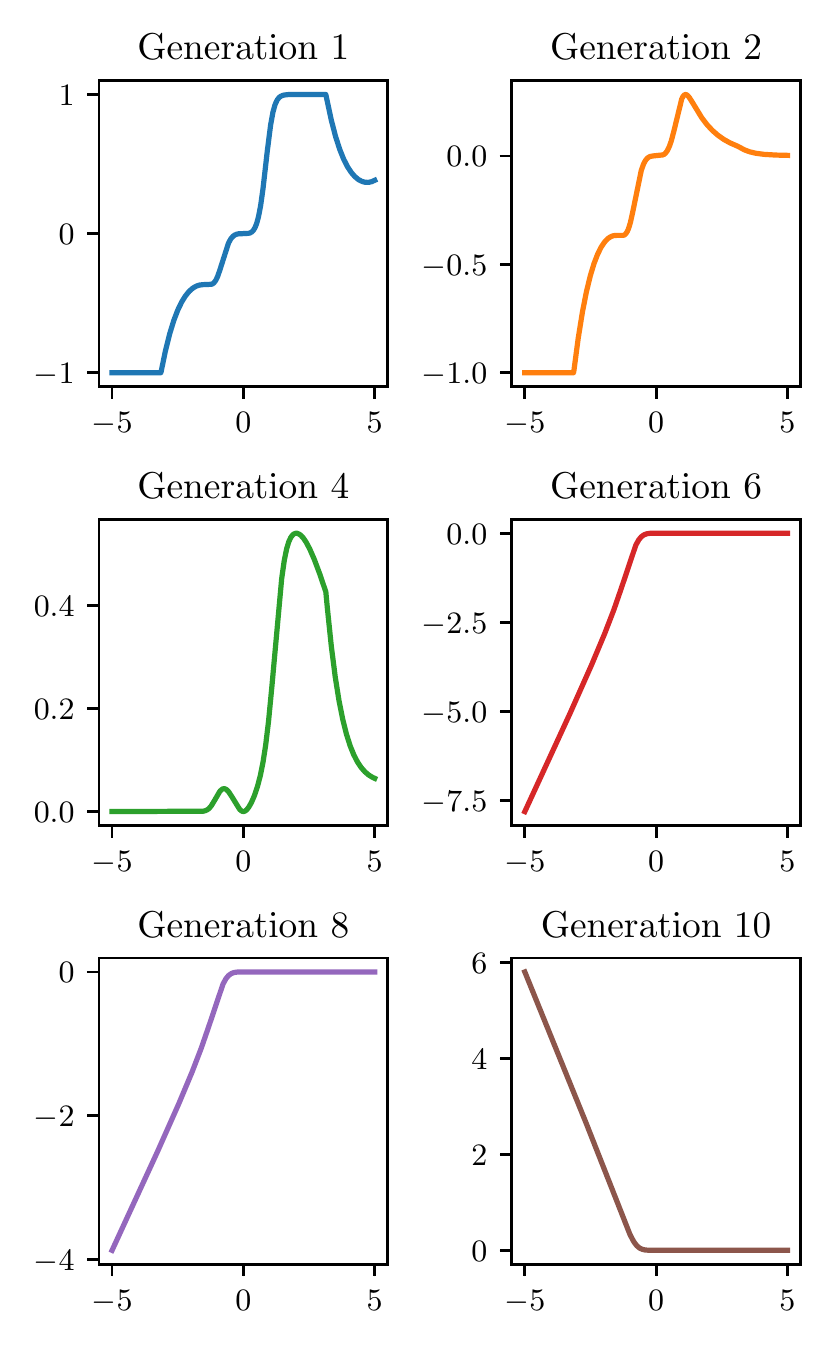}
    \caption{Best activation function by WRN-28-10 validation accuracy after 50 epochs of training on CIFAR-10 using evolution with loss-based fitness.  Top validation accuracy improves from 92.2 in Generation 1 to 93.9 in Generation 10. Evolution is able to discover effective activation functions that are not likely to be discovered by hand.  In particular, the top activation function discovered by evolution is smooth everywhere, unlike ReLU, which is not smooth at $x=0$.  This difference is likely the reason for its superior performance.}
    \label{fig:bestpergeneration}
\end{figure}

\definecolor{color1}{rgb}{0.12156862745098039, 0.4666666666666667, 0.7058823529411765}
\definecolor{color2}{rgb}{1.0, 0.4980392156862745, 0.054901960784313725}
\definecolor{color3}{rgb}{0.17254901960784313, 0.6274509803921569, 0.17254901960784313}
\begin{table*}[]{\small
\begin{tabularx}{\linewidth}{clcc}
  &                                                                    & \multicolumn{2}{c}{\textbf{Accuracy}}                        \\\textbf{Function Plots}
 &                                                                    & \multicolumn{1}{c}{CIFAR-10} & \multicolumn{1}{c}{CIFAR-100} \\
 \multirow{4}{*}{\includegraphics{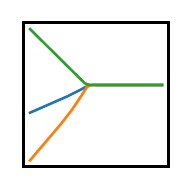}} & \multicolumn{1}{c}{\textbf{Evolution with Loss-based Fitness ($S_2$)}}                                                         &   &         \\
 & {\LARGE \textcolor{color1}{\textbullet}} $(e^{(\min\{\textrm{erf}(x), 0\}) - (\max\{x, 0\})}) * (\min\{(\arctan((x)^3)) * (\max\{|x|, 0\}), 0\})$              & \textbf{94.1} & 73.9 \\
 & {\LARGE \textcolor{color2}{\textbullet}} $(e^{\max\{\min\{\textrm{erf}(x), 0\}, \max\{x, 0\}\}}) * (\min\{(\arctan((x)^3)) * (\max\{|x|, 0\}), 0\})$         & 10.0 & 01.0 \\
 & {\LARGE \textcolor{color3}{\textbullet}} $(-((\arctan((x)^3)) * (\cos(1)))) * (-((\arctan(\min\{x, 0\})) * (\max\{|x|, 0\})))$                               & 93.9 & 74.1 \\ \\
 \multirow{4}{*}{\includegraphics{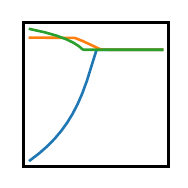}} & \multicolumn{1}{c}{\textbf{Evolution with Accuracy-based Fitness ($S_2$)}}                                                     &   &         \\
 & {\LARGE \textcolor{color1}{\textbullet}} $\min\{e^{-(\min\{(\sinh(x))^2, (0)^2\})^2}, \min\{\min\{\textrm{erf}(\log(1 + e^x)), \textrm{arcsinh}(x)\}, 0\}\}$ & 10.0 & 01.0 \\
 & {\LARGE \textcolor{color2}{\textbullet}} $\min\{\cos(\max\{(\min\{x, 0\})^3, \log(1 + e^1)\}), e^{-((|\max\{x, 0\}|) + (e^{\sigma(x)}))^2}\}$                  & 93.5 & 72.5 \\
 & {\LARGE \textcolor{color3}{\textbullet}} $\max\{\max\{(\log(|(\min\{x, 0\}) + \epsilon|)) * (\sigma(\textrm{erf}(x))), 0\}, 0\}$ \hfill (*) 
 & 92.5 & 72.3 \\ \\
 \multirow{4}{*}{\includegraphics{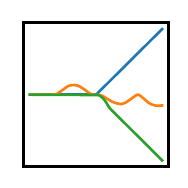}} & \multicolumn{1}{c}{\textbf{Random Search ($S_2$)}}                                                                          &   &         \\
 & {\LARGE \textcolor{color1}{\textbullet}} $(\min\{\max\{(\log(|(x) + \epsilon|))^3, -(\log(|(x) + \epsilon|))\}, 0\}) + (e^{(\min\{\tanh(x), 0\}) + (\log(|(\max\{x, 0\}) + \epsilon|))})$ \hfill (*) & 93.8 & 73.9 \\
 & {\LARGE \textcolor{color2}{\textbullet}} $(\arctan(\min\{\sinh(\sin(x)), \arctan(\max\{x, 0\})\})) * (\tanh((-(\sin(x))) * (\textrm{arcsinh}(x))))$          & 93.3 & 72.1 \\ 
 & {\LARGE \textcolor{color3}{\textbullet}} $(\max\{\frac{\min\{(x)^3, 0\}}{\max\{\sin(x), 0\} + \epsilon}, 0\}) - (\min\{(x)^2, \max\{x, 0\}\})$                          & 93.9 & 73.2 \\ \\
 \multirow{4}{*}{\includegraphics{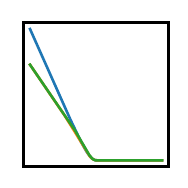}} & \multicolumn{1}{c}{\textbf{Exhaustive Search ($S_1$)}}                                                &   &         \\
 & {\LARGE \textcolor{color1}{\textbullet}} $(\arctan(x)) * (\min\{x, 0\})$                                                       & 94.0 & \textbf{74.5} \\
 & {\LARGE \textcolor{color2}{\textbullet}} $(\tanh(x)) * (\min\{x, 0\})$                                                         & \textbf{94.1} & 74.3 \\
 & {\LARGE \textcolor{color3}{\textbullet}} $(\min\{x, 0\}) * (\textrm{erf}(x))$                                                  & 94.0 & 74.2 \\ \\
 \multirow{4}{*}{\includegraphics{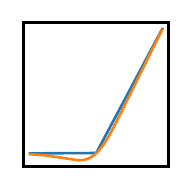}} & \multicolumn{1}{c}{\textbf{Baseline Activation Functions}}       &   &         \\
 & {\LARGE \textcolor{color1}{\textbullet}} ReLU$(x)$                                                                           & 94.0 & 73.3 \\
 & {\LARGE \textcolor{color2}{\textbullet}} Swish$(x)$                                                                          & 93.8 & 71.1\\ \\ \\ \\
\end{tabularx}}
\caption{The top three activation functions discovered by each search strategy, along with the baseline activation functions ReLU and Swish.  The functions achieved the highest validation set accuracy after 50 epochs of training with WRN-28-10 on CIFAR-10.  The final test accuracies listed on the right are the median of five runs after training WRN-28-10 from scratch for 200 epochs with each activation function on CIFAR-10 and CIFAR-100.  Function plots have domain $x \in (-5, 5)$ but have different ranges.  Functions marked with an asterisk (*) occasionally did not train to completion due to asymptotes at $x = -\epsilon$.  Exhaustive search finds multiple activation functions in a simple search space ($S_1$) that consistently outperform ReLU and Swish.  Evolution with loss-based fitness was the only technique that was able to discover such activation functions in a more complicated search space ($S_2$), suggesting that it is the most promising technique for scaling up in the future.}
\label{tab:results}
\end{table*}

\subsection{Crossover}
In crossover, two parent activation functions exchange randomly selected subtrees, producing one new child activation function.  The subtrees are constrained to be of the same depth, ensuring the child activation function is a member of the same search space as its parents.  Crossover is depicted in Figure~\ref{fig:crossover}.


\subsection{Evolution}
Starting with a population of $N$ activation functions, a neural network is trained with each function on a given training dataset.  Each function is assigned a fitness $p_i$ equal to the softmax of an evaluation metric $L_i$.  This metric could be either accuracy or negative loss obtained on the validation dataset.  More specifically, 
\begin{equation}
    p_i = \frac{e^{L_i}}{\sum\limits_{j=1..N}e^{L_j}}.
\end{equation}
The softmax operation converts the fitness values to a probability distribution, allowing functions to be randomly sampled.  From the $N$ activation functions, $2(N-m)$ are selected with replacement for reproduction with probability proportional to their fitness.  Crossover followed by mutation is applied to the selected activation functions to obtain a new population of size $N-m$.  In order to increase exploration, $m$ randomly generated functions are added to a population that will again be of size $N$.  This process is repeated for several generations, and the activation functions with the best performance over the history of the search are returned as a result.

\section{Experiments and Setup}
\label{sec:experiments}
This section presents experiments with multiple architectures, \allowbreak datasets, and search strategies.

\begin{table*}[]
    \centering
    \begin{multicols}{2}
    \begin{tabular}{ccc} \hline
         Activation Function & Mean Accuracy (95\% C.I.) & Repeats \\ \hline
         Best from $S_1$ & 74.2 ($\pm 0.1$) & 50 \\
         Best from $S_2$ & 74.0 ($\pm 0.2$) & 15 \\
         ReLU            & 73.2 ($\pm 0.2$) & 50 \\
         Swish           & 49.6 ($\pm 11.6$) & 25 \\ \hline \\
    \end{tabular}
    
    \begin{tabular}{cc} \hline
    Activation Functions & $t$-statistic; $p$-value \\ \hline
    Best from $S_1$ vs. ReLU  & 9.73; $4.64 \times 10^{-16}$\\
    Best from $S_1$ vs. Swish & 5.91; $9.91 \times 10^{-8}$\\
    Best from $S_2$ vs. ReLU  & 4.51; $2.91 \times 10^{-5}$\\
    Best from $S_2$ vs. Swish & 3.17; $2.98 \times 10^{-3}$\\ \hline \\
    \end{tabular}
    \end{multicols}
    
    \caption{Confidence intervals (95\%) and independent $t$-tests comparing mean accuracies after training WRN-28-10 on CIFAR-100 for 200 epochs with the best function from $S_1$, $(\arctan(x)) * (\min\{x, 0\})$, the best function from $S_2$, $(-((\arctan((x)^3)) * (\cos(1)))) * (-((\arctan(\min\{x, 0\})) * (\max\{|x|, 0\})))$, and baseline functions ReLU and Swish.  The discovered functions perform significantly better than both baselines.  WRN-28-10 was not trained 50 times with Best from $S_2$ and with Swish due to time constraints.  In repeated trials, Swish occasionally caused the network to stall during training, explaining its low mean accuracy.}
    
    \label{tab:statistical_significance}
\end{table*}


\subsection{Architectures and Datasets}

A wide residual network \cite{zagoruyko2016wide} with depth 28 and widening factor 10 (WRN-28-10), implemented in TensorFlow \cite{abadi2016tensorflow}, is trained on the CIFAR-10 and CIFAR-100 image datasets \cite{krizhevsky2009learning}.  The architecture is comprised of repeated residual blocks which apply batch normalization and ReLU prior to each convolution.  In the experiments, all ReLU activations are replaced with a candidate activation function.  No other changes to the architecture are made. Hyperparameters are chosen to mirror those of Zagoruyko and Komodakis \cite{zagoruyko2016wide} as closely as possible.  Featurewise center, horizontal flip, and ZCA whitening preprocessing are applied to the datasets.  Dropout probability is 0.3, and the architecture is optimized using stochastic gradient descent with Nesterov momentum 0.9.  WRN-40-4 (a deeper and thinner wide residual network architecture) is also used in some experiments for comparison.

The CIFAR-10 and CIFAR-100 datasets both have 50K training images, 10K testing images, and no standard validation set.  To prevent overfitting, balanced validation splits are created for both datasets by randomly selecting 500 images per class from the CIFAR-10 training set and 50 images per class from the CIFAR-100 training set.  The test set is not modified so that the results can be compared with other work.

  
To discover activation functions, a number of search strategies are used.  Regardless of the strategy, the training set always consists of 45K images while the validation set contains 5K images; the test set is never used during the search.  All ReLU activations in WRN-28-10 are replaced with a candidate activation function and the architecture is trained for 50 epochs.  The initial learning rate is set to 0.1, and decreases by a factor of 0.2 after epochs 25, 40, and 45.  Training for only 50 epochs makes it possible to evaluate many activation functions without excessive computational cost.

The top three activation functions by validation accuracy from the entire search are returned as a result.  For each of these functions, a WRN-28-10 is trained from scratch for 200 epochs.  The initial learning rate is set to 0.1, and decreases by a factor of 0.2 after epochs 60, 120, and 160, mirroring the work by Zagoruyko and Komodakis \cite{zagoruyko2016wide}.  After training is complete, the test set accuracy is measured.  The median test accuracy of five runs is reported as the final result, as is commonly done in similar work in the literature \cite{ramachandran2017searching}.

\subsection{Search Strategies}
Three different techniques are used to explore the space of activation functions: exhaustive search, random search, and evolution. Exhaustive search evaluates every function in $S_1$, while random search and evolution explore $S_2$. It is noteworthy that evolution is able to discover high-performing activation functions in $S_2$, where the search space contains over 41 billion possible function strings.    

\subsubsection{Exhaustive Search}

Ramachandran et al. search for activation functions using reinforcement learning and argue that simple activation functions consistently outperform more complicated ones \cite{ramachandran2017searching}.  Although evolution is capable of discovering high-performing, complex activation functions in an enormous search space, exhaustive search can be effective in smaller search spaces.  $S_1$, for example, contains 3,456 possible function strings.
\subsubsection{Random Search}

An illustrative baseline comparison with evolution is random search.  Instead of evolving a population of 50 activation functions for 10 generations, 500 random activation functions from $S_2$ are grouped into 10 ``generations'' of 50 functions each.

\subsubsection{Evolution}
As shown in Figure \ref{fig:evolution_max}, evolution discovers better activation functions more quickly than random search in $S_2$, a search space where exhaustive search is infeasible.  During evolution, candidate activation functions are assigned a fitness value based on either accuracy or loss on the validation set.  Accuracy-based fitness favors exploration over exploitation: activation functions with poor validation accuracy still have a reasonable probability of surviving to the next generation.  A hypothetical activation function that achieves 90\% validation accuracy is only 2.2 times more likely to be chosen for the next generation than a function with only 10\% validation accuracy since $e^{0.9} / e^{0.1} \approx 2.2$.

Loss-based fitness sharply penalizes poor activation functions.  It finds high-performing activation functions more quickly, and gives them greater influence over future generations.  An activation function with 0.01 validation loss is 21,807 times more likely to be selected for the following generation than a function with a validation loss of 10.  $e^{-0.01} / e^{-10} \approx 21807$.


Both experiments begin with an initial population of 50 random activation functions $(N=50)$, and run through 10 generations of evolution.  Each new generation of 50 activation functions is comprised of the top five functions from the previous generation, 10 random functions $(m=10)$, and 35 functions created by applying crossover and mutation to existing functions in the population, as described in Section \ref{sec:evolution}. Ten generations of evolution takes approximately 2,000 GPU hours using GeForce GTX 1080 GPUs.

\subsection{Activation Function Specialization}

An important question is the extent to which activation functions are specialized for the architecture and dataset for which they were evolved, or perform well across different architectures and datasets.  To address this question, activation functions discovered for WRN-28-10 on CIFAR-10 are transferred to WRN-40-4 on CIFAR-100.  These activation functions are compared with the best from a small search (1.9K activation functions from $S_1$) with WRN-40-4 on CIFAR-100.

\section{Results}
\label{sec:results}
This section presents the experimental results, which demonstrate that evolved activation functions can outperform baseline functions like ReLU and Swish.

\begin{table}[]
    \centering
    \begin{tabular}{cc} \hline
        Activation Function & Accuracy \\ \hline
        $\sigma(x) * \textrm{erf}(x)$ & \textbf{72.6} \\
        $\tanh(x) * \min\{x, 0\}$ & 72.1 \\
        $\textrm{Swish}(x)$ & 71.1 \\
        $\textrm{ReLU}(x)$ & 71.0 \\ \hline \\
    \end{tabular}
    \caption{Test set accuracy of WRN-40-4 with various activation functions after 200 epochs of training on CIFAR-100.  Results reported are median of five runs.  The top activation function discovered for WRN-28-10 on CIFAR-10, $\tanh(x) * \min\{x, 0\}$, successfully transfers to this new task, outperforming both baselines.  However, a search designed specifically for WRN-40-4 on CIFAR-100 discovers a novel activation function, {\normalfont $\sigma(x) * \textrm{erf}(x)$} that results in even higher performance.  This result demonstrates that the main power of activation function metalearning is to be able to specialize the function to the architecture and dataset.}
    \label{tab:specialized}
\end{table}

\subsection{Improving Performance}

Table \ref{tab:results} lists the activation functions that achieved the highest validation set accuracies after 50 epochs of training with WRN-28-10 on CIFAR-10.  The top three activation functions for each search strategy are included.  To emulate their true performance, a WRN-28-10 with each activation function was trained for 200 epochs five times on both CIFAR-10 and CIFAR-100 and evaluated on the test set.  The median accuracy of these tests is reported in Table~\ref{tab:results}.  Although no search was performed on CIFAR-100 with WRN-28-10, the functions that perform well on CIFAR-10 successfully generalize to CIFAR-100.

The best three activation functions discovered through exhaustive search in $S_1$ outperform ReLU and Swish. This finding shows how important it is to have an effective search method. There are good functions even in $S_1$. It is likely that there are even better functions in $S_2$, but with billions of possible functions, a more sophisticated search method is necessary.

The activation functions discovered by random search have unintuitive shapes (Table~\ref{tab:results}).  Although they fail to outperform the baseline activation functions, it is impressive that they still consistently reach a reasonable accuracy.  One of the functions (marked with (*) in Table \ref{tab:results}) discovered by random search occasionally failed to train to completion due to an asymptote at $x = -\epsilon$.

Evolution with accuracy-based fitness is less effective because it does not penalize poor activation functions severely enough.  One of the functions failed to learn anything better than random guessing.  It was likely too sensitive to random initialization or was unable to learn with the slightly different learning rate schedule of a full 200-epoch training.  Another function (marked with (*) in Table \ref{tab:results}) often did not train to completion due to an asymptote at $x = -\epsilon$.  The one function that consistently trained well still failed to outperform ReLU.


Evolution with loss-based fitness is able to find good functions in $S_2$. One of the three activation functions discovered by evolution outperformed both ReLU and Swish on CIFAR-10, and two of the three discovered outperformed ReLU and Swish on CIFAR-100.  Figure \ref{fig:bestpergeneration} shows the top activation function after each generation of loss-based evolution.  This approach discovered both novel and unintuitive functions that perform reasonably well, as well as simple, smooth, and monotonic functions that outperform ReLU and Swish. It is therefore the most promising search method in large spaces of activation functions.

\begin{figure}
    \centering
    \includegraphics{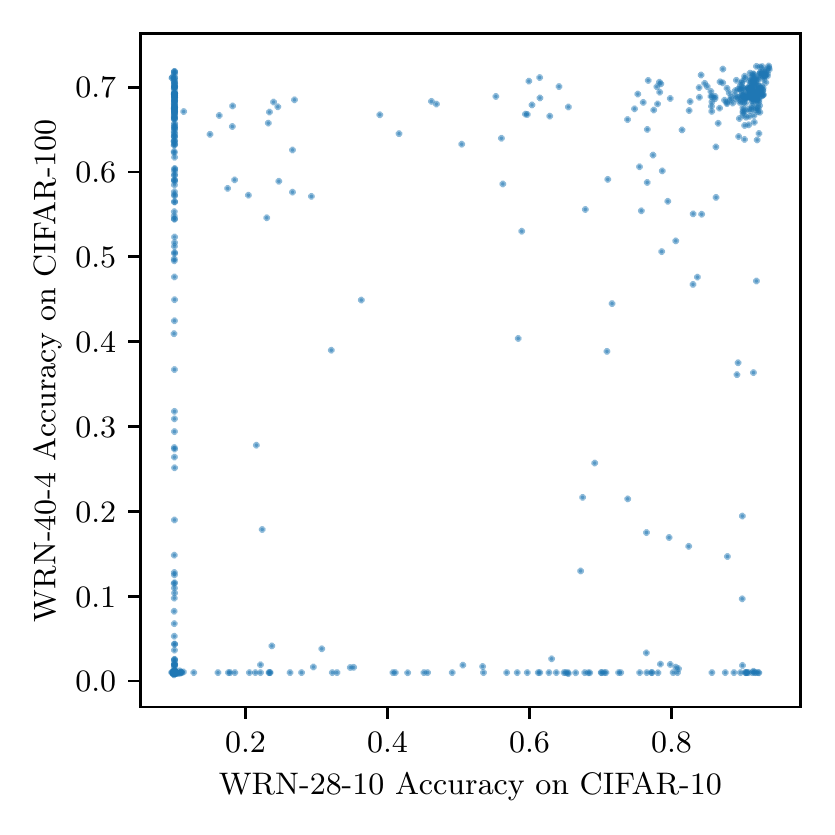}
    \caption{Activation function accuracy across tasks.  Each data point represents validation accuracy when training with a given activation function from $S_1$ with WRN-40-4 on CIFAR-100 for 50 epochs, and with WRN-28-10 on CIFAR-10 for 50 epochs.  Some activation functions perform well when paired with a different architecture and dataset.  Other functions are specialized to a given architecture and dataset, and do not transfer.  The results suggest that reasonable performance can be expected from general evolved activation functions, but that the best performance comes from evolving activation functions for a specific task.}
    \label{fig:scatter}
\end{figure}

\begin{figure*}
    \centering
    \includegraphics{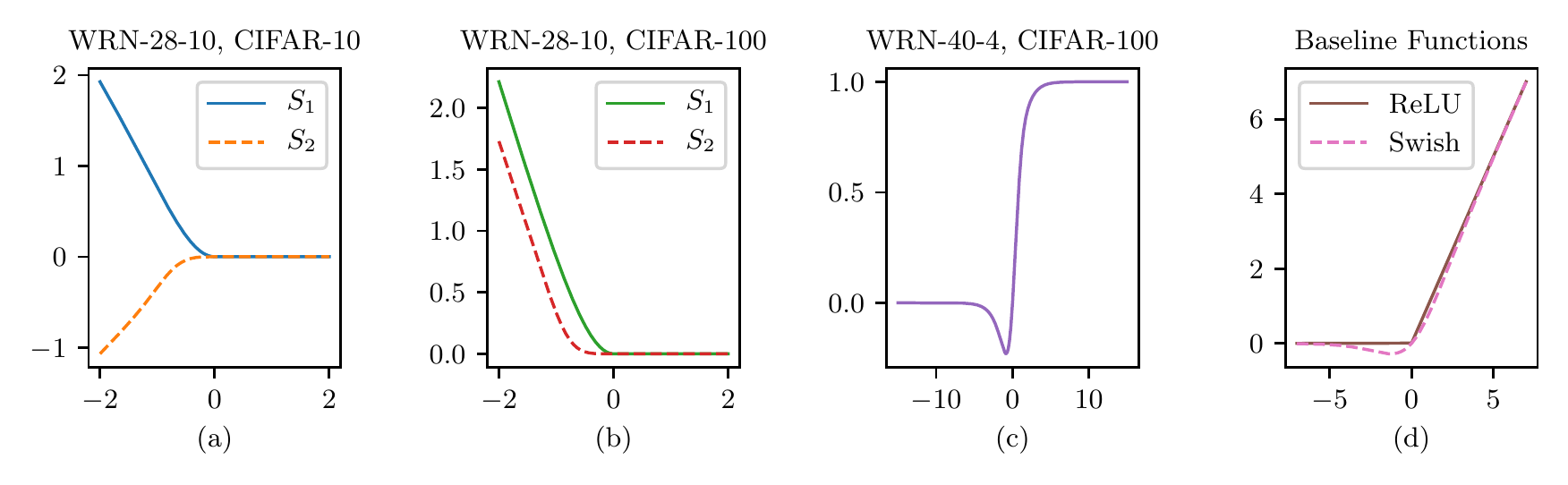}
    \caption{A summary of the best activation functions found in this paper.  (a): Exhaustive search in $S_1$ discovers $(\tanh(x)) * (\min\{x, 0\})$, and evolution in $S_2$ discovers {\normalfont $(e^{(\min\{\textrm{erf}(x), 0\}) - (\max\{x, 0\})}) * (\min\{(\arctan((x)^3)) * (\max\{|x|, 0\}), 0\})$}.  Both functions achieve a median test accuracy of 94.1 with WRN-28-10 on CIFAR-10, outperforming that of ReLU (94.0) and Swish (93.8).  (b): The functions $(\arctan(x)) * (\min\{x, 0\})$ in $S_1$ and $(-((\arctan((x)^3)) * (\cos(1)))) * (-((\arctan(\min\{x, 0\})) * (\max\{|x|, 0\})))$ in $S_2$ outperform ReLU and Swish by statistically significant margins with WRN-28-10 on CIFAR-100, demonstrating the power of novel activation functions.  (c): Functions evolved for WRN-28-10 on CIFAR-10 perform well with WRN-40-4 on CIFAR-100, but a new function discovered specifically for WRN-40-4 on CIFAR-100, {\normalfont $\sigma(x) * \textrm{erf}(x)$}, achieves even higher performance.  This result shows that the biggest advantage of activation function metalearning is the ability to discover functions that are specialized to the architecture and dataset.  (d): The baseline activation functions, ReLU and Swish, are included for visual comparison.}
    \label{fig:summary}
\end{figure*}

The performance gains on CIFAR-10 are consistent but small, and the improvement on CIFAR-100 is larger.  It is possible that more difficult datasets provide more room for improvement that a novel activation function can exploit.


To evaluate the significance of these results, WRN-28-10 was trained on CIFAR-100 for 200 epochs 50 times with ReLU, 50 times with the best function found by exhaustive search in $S_1$, $(\arctan(x)) * (\min\{x, 0\})$, 25 times with Swish, and 15 times with the best function found by evolution in $S_2$, $(-((\arctan((x)^3)) * (\cos(1)))) * \phantom{}$ \break $(-((\arctan(\min\{x, 0\})) * (\max\{|x|, 0\})))$.  Table \ref{tab:statistical_significance} shows 95\% confidence intervals and the results of independent $t$-tests comparing the mean accuracies achieved with each activation function.  The results show that replacing a baseline activation function with an evolved one results in a statistically significant increase in accuracy.

\subsection{Specialized Activation Functions}
\label{sec:specialized}
Since different functions are seen to emerge in different experiments, an important question is: How general or specialized are they to a particular architecture and dataset?  To answer this question, the top activation function discovered for WRN-28-10 on CIFAR-10, $\tanh(x) * \min\{x, 0\}$, was trained with WRN-40-4 on CIFAR-100 for 200 epochs.  This result was then compared with performance achieved by $\sigma(x) * \textrm{erf}(x)$, an activation function discovered specifically for WRN-40-4 on CIFAR-100.  Table \ref{tab:specialized} summarizes the result: The activation function discovered for the first task does transfer to the second task, but even higher performance is achieved when a specialized function is discovered specifically for the second task.

The specialized activation function, $\sigma(x) * \textrm{erf}(x)$, is shown in Figure \ref{fig:summary}c.  It is similar to $\sigma(x)$ in that it tends towards 0 as $x \rightarrow -\infty$, and approaches 1 as $x \rightarrow \infty$.  It differs from $\sigma(x)$ in that it has a non-monotonic bump for small, negative values of $x$.  A 50-epoch training of WRN-40-4 on CIFAR-100 with activation $\sigma(x)$ achieved validation accuracy of just 63.2.  The superior performance of $\sigma(x) * \textrm{erf}(x)$ suggests that the negative bump was important, as the shapes of the two activation functions are otherwise similar. This result demonstrates how evolution can discover specializations that make a significant difference.


\section{Discussion and Future Work}
\label{sec:discussion}

Among the top activation functions discovered, many are smooth and monotonic.  Hand-engineered activation functions frequently share these properties \cite{nwankpa2018activation}.  Two notable exceptions were found by random search and evolution with accuracy-based fitness.  Although these functions do not outperform ReLU, the fact that WRN-28-10 was able to achieve such high accuracy with these arbitrary functions raises questions as to what makes an activation function effective. Ramachandran et al.~\cite{ramachandran2017searching} asserted that simpler activation functions consistently outperformed more complicated ones. However, the high accuracy achieved with activation functions discovered by evolution in $S_2$ demonstrates that complicated activation functions can compete with simpler ones. Such flexibility may be particularly useful in specialization to different architectures and datasets. It is plausible that there exist many unintuitive activation functions which can outperform the more general ones in specialized settings.  Evolution is well-positioned to discover them.

Activation functions discovered by evolution perform best on the architectures and datasets for which they were evolved.  Figure \ref{fig:scatter} demonstrates this principle.  More generally, it shows the performance of several activation functions when trained with WRN-28-10 for 50 epochs on CIFAR-10 and when trained with WRN-40-4 for 50 epochs on CIFAR-100.  Activation functions that perform well for one task often perform well on another task, but not always.  Therefore, if possible, one should evolve them specifically for each architecture and dataset.  However, as the results in Section \ref{sec:results} show, it is feasible to evolve using smaller architectures and datasets and then transfer to scaled up architectures and more difficult datasets within the same domain.

In the future, it may be possible to push such generalization further, by evaluating functions across multiple architectures and datasets.  In this manner, evolution may be able to combine the requirements of multiple tasks, and discover functions that perform well in general.  However, the main power in activation function metalearning is to discover functions that are specialized to each architecture and dataset.  In that setting most significant improvements are possible.


\section{Conclusion}
\label{sec:conclusion}

Multiple strategies for discovering novel, high-performing activation functions were presented and evaluated: namely exhaustive search in a small search space ($S_1$) and random search and evolution in a larger search space ($S_2$).  Evolution with loss-based fitness finds activation functions that achieve high accuracy and outperform standard functions such as ReLU and novel functions such as Swish, demonstrating the power of search in large spaces. The best activation functions successfully transfer from CIFAR-10 to CIFAR-100 and from WRN-28-10 to WRN-40-4. However, the main power of activation function metalearning is in finding specialized functions for each architecture and dataset, leading to significant improvement.

  

\begin{acks}
    This research was supported in part by NSF under grant DBI-0939454 and by DARPA under grant FA8750-18-C-0103.
    The authors acknowledge the Texas Advanced Computing Center (TACC) at The University of Texas at Austin for providing HPC resources that have contributed to the research results reported within this paper.
    

\end{acks}

\bibliographystyle{ACM-Reference-Format}
\bibliography{acmart} 

\end{document}